\documentclass[letterpaper]{article} 
\usepackage{aaai25}  
\usepackage{times}  
\usepackage{helvet}  
\usepackage{courier}  
\usepackage[hyphens]{url}  
\usepackage{graphicx} 
\usepackage{natbib}  
\usepackage{caption} 
\usepackage{algorithm}
\usepackage{algorithmic}
\usepackage{booktabs}
\usepackage{multirow}
\usepackage{newfloat}
\usepackage{listings}
\DeclareCaptionStyle{ruled}{labelfont=normalfont,labelsep=colon,strut=off} 
\floatstyle{ruled}
\newfloat{listing}{tb}{lst}{}
\floatname{listing}{Listing}
\title{Mitigating Language Bias in Cross-Lingual Job Retrieval: A Recruitment Platform Perspective}
\author{
    Napat Laosaengpha\textsuperscript{\rm 1}, Thanit Tativannarat\textsuperscript{\rm 1}, Attapol Rutherford\textsuperscript{\rm 2}, Ekapol Chuangsuwanich\textsuperscript{\rm 1}
}
\affiliations{
    \textsuperscript{\rm 1} Department of Computer Engineering, Faculty of Engineering, Chulalongkorn University \\
    \textsuperscript{\rm 2} Department of Linguistics, Faculty of Arts, Chulalongkorn University \\
    \small{\texttt{ \{napatnicky,thanit.tati\}@gmail.com \{attapol.t,Ekapol.C\}@chula.ac.th} }
    
}

\usepackage{bibentry}

\begin{document}

\maketitle

\begin{abstract}
Understanding the textual components of resumes and job postings is critical for improving job-matching accuracy and optimizing job search systems in online recruitment platforms. However, existing works primarily focus on analyzing individual components within this information, requiring multiple specialized tools to analyze each aspect. Such disjointed methods could potentially hinder overall generalizability in recruitment-related text processing. Therefore, we propose a unified sentence encoder that utilized multi-task dual-encoder framework for jointly learning multiple component into the unified sentence encoder. The results show that our method outperforms other state-of-the-art models, despite its smaller model size. Moreover, we propose a novel metric, Language Bias Kullback–Leibler Divergence (LBKL), to evaluate language bias in the encoder, demonstrating significant bias reduction and superior cross-lingual performance.
\end{abstract}

%

\section{Introduction}

The online job recruitment platforms have emerged as essential tools to streamline and accelerate the talent acquisition process. These platforms usually rely on essential details from resume and job posting to facilitate the connection between recruiters and job seekers. The resumes showcase personal overview of candidate's capabilities including work experience, skills and expertise, whereas the job postings outline job title, specialties and responsibility for open positions. 

Therefore, understanding the semantic meaning of the textual information within resumes and job postings would greatly facilitate the matchmaking process on the online job recruitment platform, where the underlying process is an automatic job recommendation system \cite{DBLP:journals/corr/abs-2107-00221}. Additionally, this understanding could be further developed into analytical tools for various job-related tasks such as job mobility prediction \cite{zha2024towards} and job demand forecasting \cite{lu2022human}.


Previous approaches to understanding this textual information as low-dimensional dense sentence representations primarily focus on a individual component within the resumes and the job posting such as job title (JT) \cite{DBLP:journals/corr/abs-2109-09605,laosaengpha-etal-2024-learning}, job description (JD) \cite{goyal-etal-2023-jobxmlc} and skills set \cite{lin2023skill}. 


However, learning sentence representations in these previous works are constrained to handling the individual component, which limit their versatility in being applied to analyze the other job-related components. Moreover, the existing works often prioritize the study on a mainstream language like English, leaving non-English language especially in low-resource ones under-explored. It might be due to the scarcity of public dataset and the specialized nature of the job recruitment domain for non-English language. Additionally, their studies mainly cover monolingual setting, which might not applied to regions where the users often alternate between languages on the platform.

In this paper, we propose a multi-task learning framework to develop a bilingual sentence encoder (Thai and English) for general-purpose use in the recruitment domain by using label-free information from online user-generated job postings. We leverage a multi-task dual encoder framework, which is simultaneously trained on three proposed job-related tasks. The job-related training tasks consist of three tasks: (A) job title translation ranking, (B) job description-title matching, and (C) job field classification. This mitigates the aforementioned limitations by learning multiple components in a unified encoder and addressing the scarcity of human-labeled training data in low-resource language. Moreover, we conduct an comprehensive study to investigate cross-lingual capabilities and also propose a novel evaluation metric to quantify language bias in the bilingual sentence encoder, where this study could be linked to application in evaluating job search systems on the platform.

Our contributions are as follows:
\begin{itemize}

    \item \textbf{Multi-task learning framework}: We propose a multi-task dual-encoder framework that jointly learns multiple components within a bilingual sentence encoder (Thai and English) for the recruitment domain. 
    
    \item \textbf{Cross-lingual and language bias analysis}: We present a novel metric to quantify language bias hidden in the sentence encoder for retrieval evaluation. We also conduct extensive study to investigate cross-lingual capabilities. The result show that our encoder achieve significantly lower language bias and improved cross-lingual performance compared to other state-of-the-art encoders.
\end{itemize}

\section{Background and Related Work}

\subsubsection{Pretraining on the Job Recruitment Domain} Several works have studied pretrained language models in the job recruitment domain. \citeauthor{qin2018enhancing} \shortcite{qin2018enhancing} and \citeauthor{cao2024tarot} \shortcite{cao2024tarot} utilized user activity data from recruitment platforms to enhance language model performance for job recommendation. \citeauthor{zhang-etal-2023-escoxlm} \shortcite{zhang-etal-2023-escoxlm} is a recent effort in multilingual pretraining across 16 European languages, leveraging ESCO taxonomy as a training data for the language model. However, this data require human-labeled training data, which is challenging to obtain for low-resource language. 

To the best of our knowledge, \citeauthor{fang2023recruitpro} \shortcite{fang2023recruitpro} is the only work that has attempted to pretrain a language model by using multiple information on job-related domain. However, this work focus on a monolingual evaluation in English. The effectiveness of the model in a bilingual setting remains unexplored, particularly in terms of cross-lingual capability and language bias in sentence representation. 


\subsubsection{Language Bias} \citeauthor{roy-etal-2020-lareqa} \shortcite{roy-etal-2020-lareqa} studied the language bias problem inside the sentence representations on a multilingual question-answering retrieval task. They found that the language bias influences performance, as queries often prefer candidates from the corresponding language while neglecting their semantic meaning. The following works proposed various methods to mitigate this language bias on the representation through post-processing matrix transformations \cite{yang-etal-2021-simple,xie-etal-2022-discovering}. Furthermore, the language bias problem also occurs in coding language embeddings \cite{utpala-etal-2024-language}. However, there is a lack of evaluation metrics to quantify the amount of language bias hidden in representations, especially in retrieval settings.

\section{Proposed Method}

 \begin{figure}[t]
\centering
\includegraphics[width=1\columnwidth]{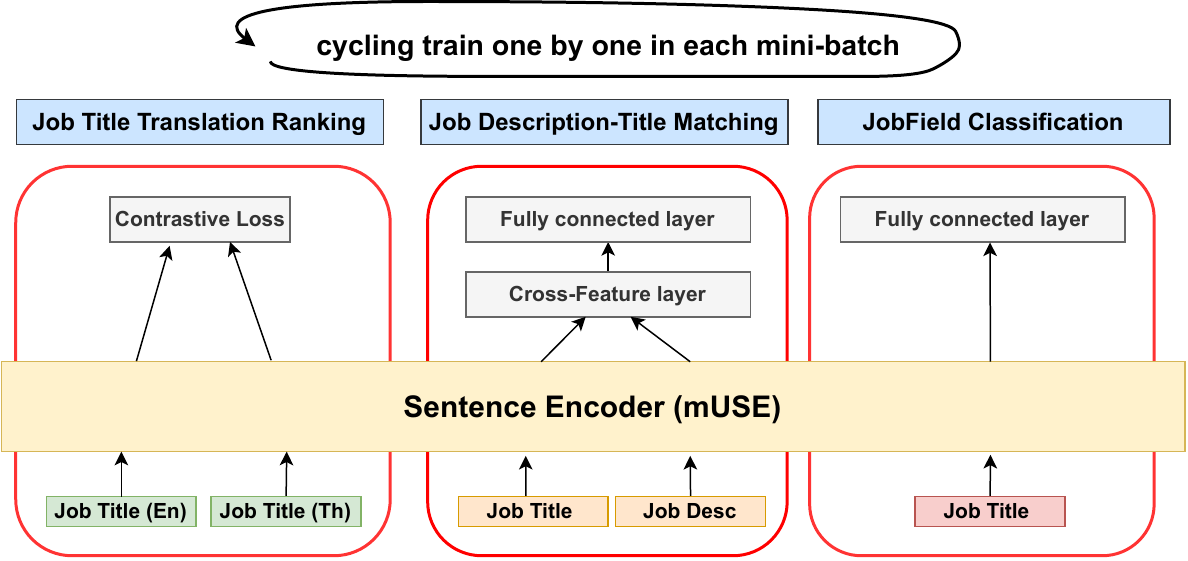}
\caption{The overview of our proposed multi-task dual-encoder framework used to train our sentence encoder. It illustrate the job title translation ranking task on the left, job description-title matching in the middle, and job field classification on the right.} 
\label{fig:first-arch}
\end{figure}

\subsection{Architecture}
An overview of our proposed method is shown in Figure~\ref{fig:first-arch}. The training process consists of three main tasks: (A) job title translation ranking, (B) job description-title matching, and (C) job field classification. Our inspiration originates from the multi-task dual encoder used in mUSE \cite{yang-etal-2020-multilingual}. However, job-related data doesn't lend itself to the traditional training tasks as described in the original paper. Therefore, we adapt to our specific requirements and designed job-related tasks to fine-tune the model. The training process is to consecutively train the model through the three job-related tasks one by one in each mini-batch, with equal weight penalty for each task.


\subsubsection{Job Title Translation Ranking Task (JT)}


The job title translation ranking task is responsible for aligning pairs of job titles that have the same semantic meaning but are in different languages to be closer in the high-dimensional space. The job title representation of both Thai and English are aligned by using contrastive loss \cite{gao-etal-2021-simcse} as a training objective to maximize a pairwise similarity between job title representations in Thai and English.
\begin{equation} 
    L_{i} = \log \frac{e^{sim(t_i,f_i)/\tau}}{\sum^N_{j=1}e^{sim(t_i,f_j)/\tau}}
\end{equation}
where $t_i$ and $f_i$ are the sentence embeddings of English and Thai job titles respectively, $sim(\cdot,\cdot)$ is the cosine similarity function, $\tau$ is the temperature scaling parameter, and $N$ is the number of negative samples.

\subsubsection{Job Description-Title Matching (JD)}

The job description and title matching is designed to predict the correlation between the job description and job title whether they have a positive or negative relationship. The task mimics the matchmaking process in job recommendation system, where the job description is used to search for the most relevant candidate job title. We adopt the architectural design from Neural Language Inference (NLI) \cite{DBLP:conf/emnlp/ConneauKSBB17} to fit our task. The criteria for determining a positive sample between a job description and title is to use a pair of these from the same job posting. In contrast, negative pairs are created by sampling the other job postings and comparing their intersection over union (IoU) score of the job fields from the job postings. Pairs with IoU values below a threshold of 0.5 are classified as negative samples.


\begin{equation} 
    IoU_{ij}  = \frac{|JF_{i} \cap JF_{j}|}{|JF_{i} \cup JF_{j}|}
\end{equation}

where $JF$ is the list of job field in the job posting. 





\subsubsection{Job Field Classification (JF)}


The job field classification is to classify the job title into multiple specified job field from a total of 28 categories. This task also demonstrates one of the interesting aspects that a job title is generally involved in more than one related job field. For instance, ``Sales engineer`` can be categorized into the field of "Sales" and "Engineer". This relationship between the job field and the job title would afford the model the benefit of enhancing its robustness in representing job titles, as it requires additional attention to handle the ambiguity inherent in the job titles.

\begin{table*}[t]
\centering
\resizebox{\textwidth}{!}{%
\begin{tabular}{@{}lccccccc@{}}
\hline
\toprule
\multicolumn{1}{c}{\multirow{2}{*}{\textbf{Method}}} & \multicolumn{1}{l}{\multirow{2}{*}{\textbf{\#Param \& Runtime}}} & \multicolumn{3}{c}{\textbf{JTG-Synonym}} & \multicolumn{3}{c}{\textbf{JTG-Occupation}} \\ \cmidrule(l){3-8} 
\multicolumn{1}{c}{}                                 & \multicolumn{1}{l}{}                                       & R@5         & R@10        & mAP@25       & Acc@1        & Acc@3        & Acc@5        \\ \midrule
XLM-R \cite{conneau-etal-2020-unsupervised}                                               & 279M / 0.46 ms                                                        & 15.11       & 18.68       & 10.27        & 47.65        & 67.33        & 76.02         \\ \midrule
LaBSE \cite{feng-etal-2022-language}                                               & 424M / 0.48 ms                                                       & 48.63       & 60.02       & 37.83        & 60.07        & 80.89        & 87.10         \\ 
BGE-m3 \cite{chen-etal-2024-m3}                                              & 567M / 0.55 ms                                                       & 49.83       & 61.11       & 38.03        & 61.22        & 82.04        & 90.06         \\ \midrule
mUSE\textsubscript{small}CNN-based \cite{yang-etal-2020-multilingual}                                                & 69M  / 0.24 ms                                                       & 
50.83         & 61.91         & 39.23          & 58.07          & 79.27          & 86.53           \\
mUSE\textsubscript{small}CNN-based (ours)                                          & 69M / 0.24 ms                                                         & \textbf{64.89}       & \textbf{79.43}       & \textbf{52.25}        & \textbf{69.53}        & \textbf{87.67}       & \textbf{92.93}         \\ \bottomrule
\end{tabular}
}
\caption{The performance evaluation of our method against other state-of-the-art models on the JTG-Synonym and JTG-Occupation evaluation datasets. The runtime complexity was measured using the JTG-Synonym dataset and evaluated on an NVIDIA RTX 3090 GPU (24GB) paired with an Intel Xeon Silver 4210 CPU (2.20GHz).}
\label{tab:main-result}
\end{table*}


\section{Experimental Setup}


We outline the evaluation dataset and method, including JTG-Synonym, JTG-Occupation, and JTG-Jobposting.


\subsubsection{JTG-Jobposting}
We use job postings from Jobtopgun.com, a renowned recruitment website in Thailand. It consists of Thai and English job titles, their descriptions, and job fields. For our framework, we used a training data set consisting of 209,785 job postings.



\subsubsection{JTG-Synonym}
The JTG-Synonym is a synonym list that includes different variants of the same job title in Thai and English. To evaluate on the JTG-Synonym, we formulated the task as a bilingual retrieval task, where the dictionary keys (job title) were used as query and all synonyms were used as the candidate pool. Each query was performed on the English and Thai candidate pools separately to calculate the R@5, R@10, and mAP@25. The final metric values were obtained by micro-averaging across every query. The test set contains 4,420 queries (2,261 in Thai, 2,103 in English, and 56 code-switching) and a candidate pool of 34,589 entries, with 16,905 in Thai and 17,684 in English.



\subsubsection{JTG-Occupation}
The JTG-Occupation is a collection of job title along with their corresponding occupation groups. The dataset contains 5,801 samples with a total of 135 unique labels. We split the data into 4,641 samples for training and 580 samples each for validation and testing. To evaluate the embeddings generated by the sentence encoder, we created a linear classifier layer on top of it. Then, we trained the classifier on top of the embedding to predict their occupation while freezing the sentence encoder.

\subsubsection{Baseline Methods}

We benchmarked our method against two categories of multilingual pretraining models: a Mask Language Model (MLM) based model - XLM-R \shortcite{conneau-etal-2020-unsupervised}, and sentence-level models - mUSE \shortcite{yang-etal-2020-multilingual}, LaBSE \shortcite{feng-etal-2022-language} and BGE-M3 \shortcite{chen-etal-2024-m3}. The base sentence encoder chosen for our method is the mUSE\textsubscript{small}CNN due to its run-time efficiency, enabling real-time processing on recruitment platforms.

\subsubsection{Implementation Details}

The temperature scaling for the job title translation ranking task is of 0.05. The top fully connected layers for both the job field classification and the job description and title matching are set as 2 dense layers with 512 dimensions. We use a batch size of 512 and uses the Adam optimizer with 3e-5 learning rate. 

\subsection{Language Bias Metric}
We propose a novel evaluation metric, \textbf{Language Bias Kullback–Leibler Divergence (LBKL)}, to measure language bias in retrieval settings. This metric compares the distribution of language proportions between the ground truth and the predicted list for each query. Given an ordered list of retrieved items, this metric is designed to assess the language bias in the retrieved list without considering model accuracy. 
\begin{equation} 
    LBKL = \frac{\sum_{i=1}^{q} \left[ P_{th}(x) \log(\frac{P_{th}(x)}{Q_{th}(x)}) +  P_{en}(x) \log(\frac{P_{en}(x)}{Q_{en}(x)}) \right]}{q}
\end{equation}

where $P_{th}(x)$ and $P_{en}(x)$ are the proportion of Thai and English in the ground truth list for each query, $Q_{th}(x)$ and $Q_{en}(x)$ are the proportion of Thai and English in the predicted list in each query, $q$ is the number of queries.



\section{Main Result}


The main results are shown in table \ref{tab:main-result}. Our method consistently outperforms all previous state-of-the-art models across all metrics in both evaluation settings on JTG-Synonym and JTG-Occupation. Furthermore, it is much smaller in terms of parameters and the runtime complexity.

\subsubsection{Ablation Study}
We perform an ablation study to explore the performance improvements achieved by each of our proposed tasks. Specifically, we evaluate the model’s performance when trained exclusively on a single task, including job title translation ranking, job description-title matching, and job field classification.
\begin{table}[ht]
\centering
\begin{tabular}{lll}
\hline
              & \multicolumn{1}{c}{\textbf{JTG-Synonym}} & \multicolumn{1}{c}{\textbf{JTG-Occupation}} \\ \hline
mUSE          & 61.91                                    & 86.53                                       \\ \hline
+ (A) JT      & 77.43 (+15.52)                           & 89.38 (+2.85)                               \\
+ (B) JD      & 66.41 (+04.50)                           & 87.77 (+1.24)                               \\
+ (C) JF      & 61.50 (- 00.41)                          & 89.20 (+2.67)                               \\ \hline
\textbf{Ours} & 79.43 (+17.52)                           & 92.93 (+6.40)                               \\ \hline
\end{tabular}%
\caption{A performance comparison for each of our proposed training tasks, evaluated using JTG-Synonym (R@10 $\uparrow$) and JTG-Occupation (Acc@5 $\uparrow$).}
\label{tab:abla-syn}
\end{table}

The results in Table \ref{tab:abla-syn} demonstrate that utilizing our multi-task dual encoder achieves superior performance compared to models trained on a single task on both settings. This suggests the model’s robustness, as it is trained to represent multiple components simultaneously in a single training step.


\section{Discussion}


\subsection{Cross-Lingual Performance}

As mentioned in the experimental setup section, we evaluated the JTG-Synonym dataset by framing it as a retrieval task, using the dictionary keys (job titles) as queries to search for all synonyms in two separate candidate pools (one containing only English and the other only Thai). This was done in order to mitigate the language bias presented in the embeddings. However, we can also extend our evaluation to a single candidate pool, combining Thai and English candidate pools into one. This approach facilitates the analysis of language bias in the models. The detailed performance for each candidate pool is reported in Table \ref{tab:multi-result}.

The result shows that the cross-lingual performance of our model significantly improves compared to XLM-R, BGE-M3, and LaBSE in both the separate and combined pool settings. However, BGE-m3 demonstrates superior results over LaBSE in the two separate pool settings, though its performance declines in the combined pool setting. This outcome will be further discussed in the following section.

\subsection{The Language Bias in Embeddings}


\subsubsection{Language Histogram}


We can further visualized the histogram of the retrieved candidates in the JTG-synonym dataset. We used the dictionary keys (job titles) as queries to calculate the similarity scores between these queries and the candidate pool (combined pool). Then, the candidate were ranked based on their similarity scores from high to low. Next, we selected the top 100 results from this ranking to count the frequency of Thai and English candidates, without considering the correctness of their corresponding labels. Finally, the counts (\#th and \#en retrieved by each query in the top-100 results) were used to plot the histogram summarizing all queries. We present the language frequency histogram, where queries are divided into three subcategories: English (EN), Thai (TH), and code-switched (CS).


 
 
 The histogram of BGE-m3 shows a bias towards the query language, implying the language bias has significantly influenced their retrieval result, where the query would prefer the candidate in the same language more than another languages. In contrast, other models such as LaBSE and mUSE, the language histogram is slightly shifted from the middle, and mUSE (ours) remains close to the midpoint.


\begin{figure}[t]
\centering

\begin{minipage}{\columnwidth}
    \centering
    \includegraphics[width=\columnwidth]{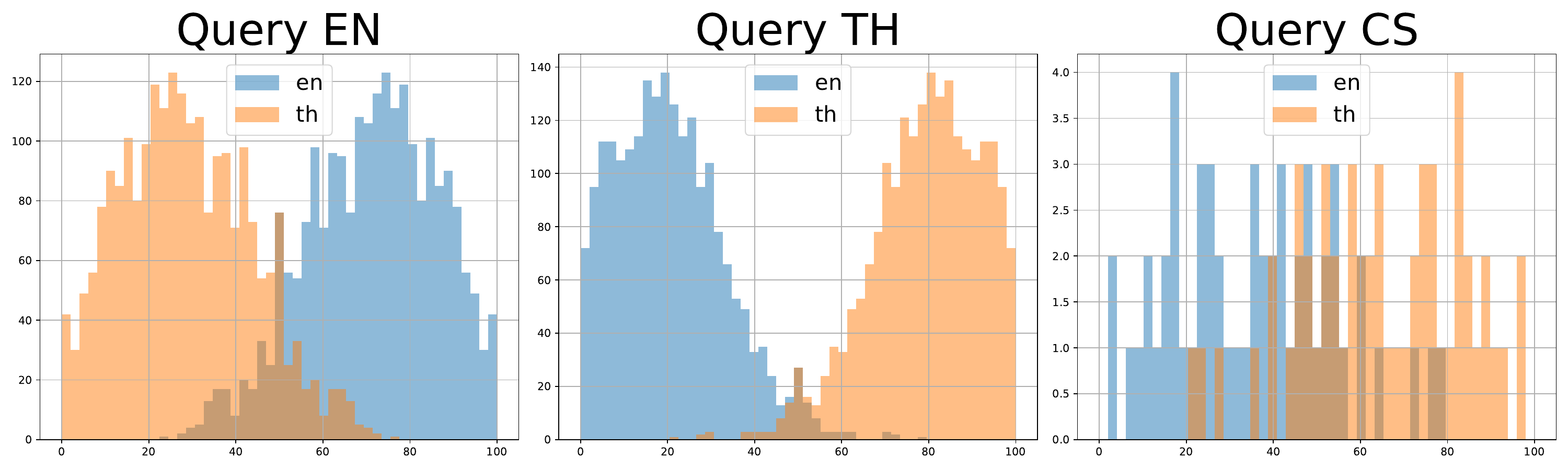}
    \caption{A language frequency histogram of LaBSE}
    \label{fig:labse-lan-bias}
\end{minipage}

\begin{minipage}{\columnwidth}
    \centering
    \includegraphics[width=\columnwidth]{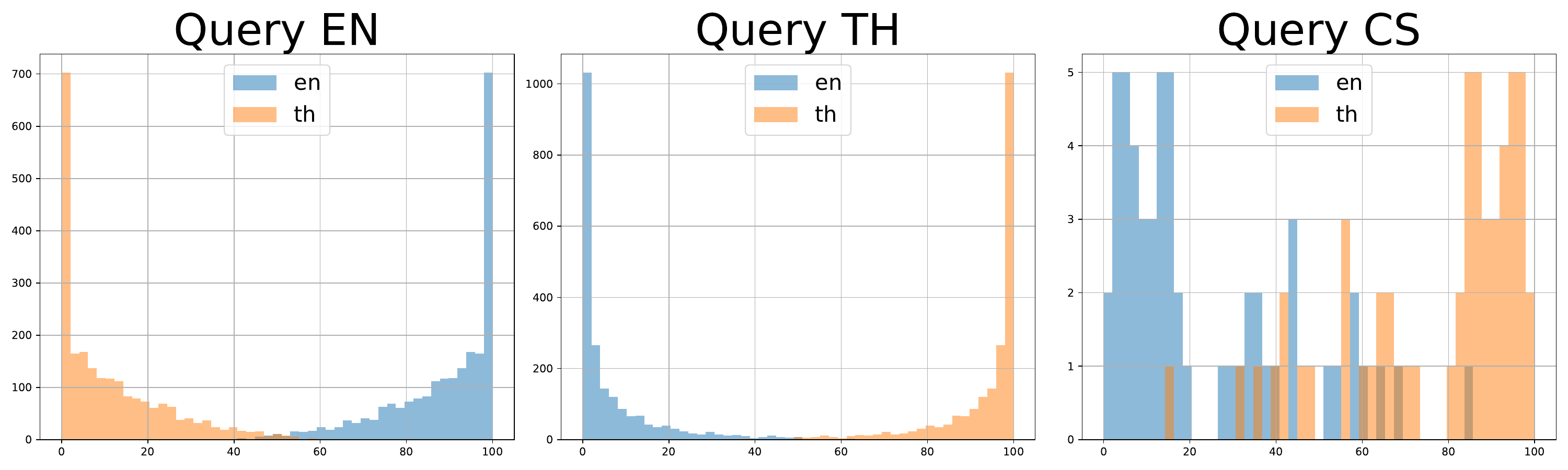}
    \caption{A language frequency histogram of BGE-M3}
    \label{fig:bge-lan-bias}
\end{minipage}


\begin{minipage}{\columnwidth}
    \centering
    \includegraphics[width=\columnwidth]{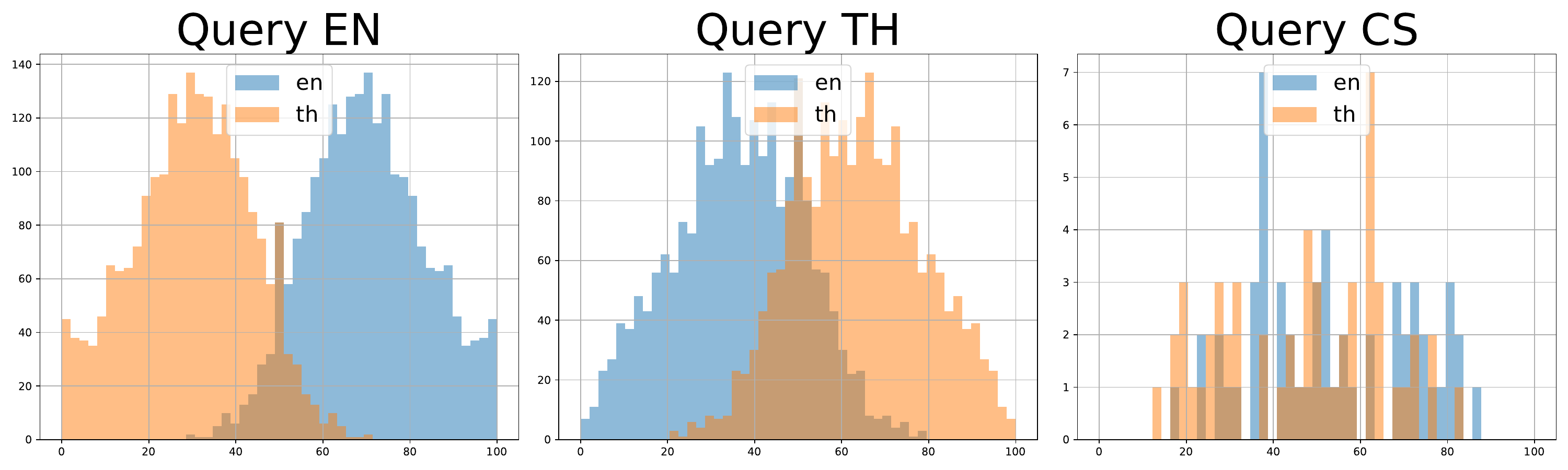}
    \caption{A language frequency histogram of mUSE}
    \label{fig:mUSE-lan-bias}
\end{minipage}

\begin{minipage}{\columnwidth}
    \centering
    \includegraphics[width=\columnwidth]{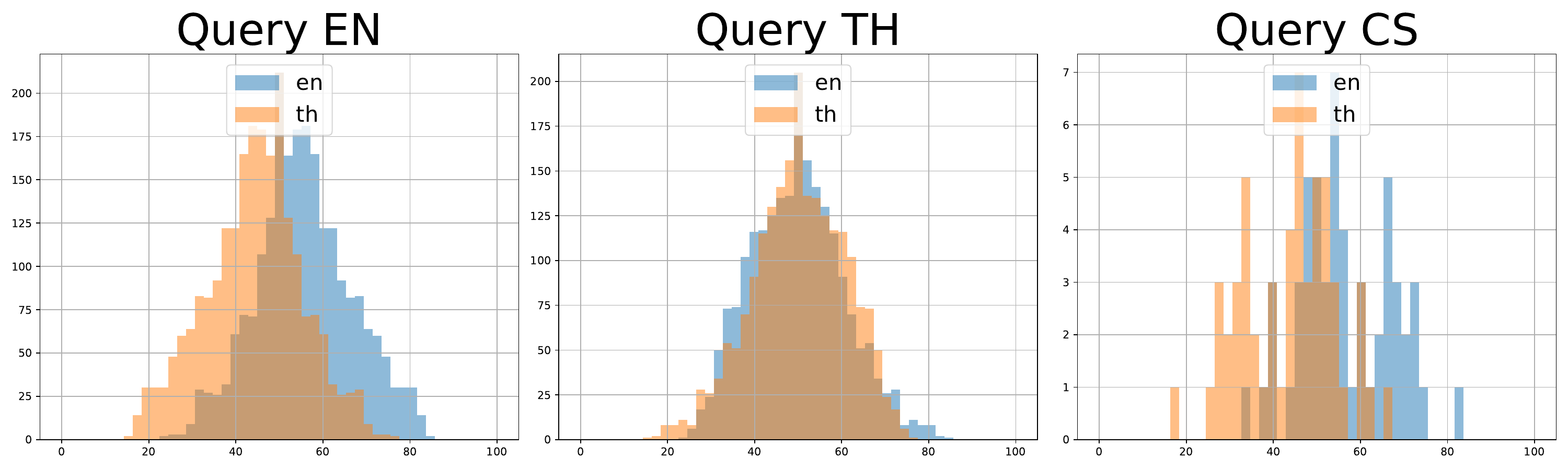}
    \caption{A language frequency histogram of mUSE (ours), with the left, middle, and right sections showing histograms for English, Thai, and code-switching queries, respectively. The orange and blue histogram represent the number of candidate results in Thai and English, respectively. }
    \label{fig:mUSE-our-lan-bias}
\end{minipage}

\end{figure}

\begin{table*}[t]
\centering
\begin{tabular}{lcccccccc|cccc}
\hline
               & \multicolumn{4}{c}{\textbf{Thai Candidate Pool}}                   & \multicolumn{4}{c|}{\textbf{English Candidate Pool}}               & \multicolumn{4}{c}{\textbf{ Combined Candidate Pool}}               \\ \hline
\textbf{Query} & \textbf{EN}    & \textbf{TH}    & \textbf{CS}    & \textit{avg}   & \textbf{EN}    & \textbf{TH}    & \textbf{CS}    & \textit{avg}   & \textbf{EN}    & \textbf{TH}    & \textbf{CS}    & \textit{avg}   \\ \hline
XLM-R          & 2.56           & 34.84          & 29.58          & 22.32 & 31.46          & 3.05           & 8.24           & 14.25 & 17.00          & 19.73          & 15.35          & 17.36 \\
LaBSE          & 59.35          & 63.97          & 57.81          & 60.37 & 59.20          & 57.84          & 67.50          & 61.51 & 45.12          & 48.62          & 48.64          & \underline{47.46} \\
BGE-m3         & 59.08          & 69.17          & 68.51          & \underline{65.58} & 62.08          & 53.30          & 78.02          & \underline{64.46} & 36.99          & 42.07          & 45.60          & 41.55 \\
\textbf{Ours}   & \textbf{82.30} & \textbf{80.18} & \textbf{82.36} & \textbf{81.61} & \textbf{75.07} & \textbf{81.18} & \textbf{83.43} & \textbf{79.89} & \textbf{67.02} & \textbf{66.82} & \textbf{59.54} & \textbf{64.46} \\ \hline
\end{tabular}%
\caption{The retrieval results (R@10 $\uparrow$) for different query-candidate-pool pairs, where the first and second columns refer to the setting of two separate pools, and the third column refers to the setting of a combined candidate pool.}
\label{tab:multi-result}
\end{table*}



\subsubsection{Language Bias Kullback–Leibler Divergence (LBKL)} 
Table \ref{tab:lan-bias-overall} presents our LBKL metrics for both in-domain (JTG-Synonym) and unseen domains namely, JTG-Skill (a list of job skills in both languages), SCB-MT (a general domain Thai-English translation dataset) \cite{lowphansirikul2022large}, and XQuAD-r (a multilingual retrieval dataset for QA, use only Thai and English corpus) \cite{roy-etal-2020-lareqa}. The experimental results on JTG-Synonym confirms that BGE-M3 and LaBSE still encounter challenges of language bias, as indicated by the language histograms and LBKL of 3.95 for BGE-M3 and 1.96 for LaBSE. This LBKL different could be linked to the cross-lingual performance in Table \ref{tab:multi-result}, as BGE-M3 outperforms in the two separate pool settings but declines in the combined pool setting.

For the unseen domains, LBKL on JTG-Skill and SCB-MT indicates a reduction in the language bias. In Xquad-r, the LBKL of our model is slightly worse than mUSE. However, the loss in performance is small. Overall, our method could potentially reduce language bias in the sentence representations across both in-domain and unseen domains. 



Table \ref{tab:abla-all} shows a language bias comparison (LBKL) for each of our proposed tasks. The JT and JD tasks reduce the language bias in most unseen domains, with only XQuAD-r showing a marginal increase. However, the LBKL of the JD task on in-domain increases slightly compared to the baseline score, and the JF task has little impact on language bias in both in-domain and unseen domains. The JT task is still crucial for reducing the language bias, consistently reducing bias in job-related domains (JTG-Synonym and JTG-Skill).

\begin{table}[ht]
\centering
\Huge
\resizebox{\columnwidth}{!}{%
\begin{tabular}{lc|ccc}
\hline
       & \textbf{In-domain} & \multicolumn{3}{c}{\textbf{Unseen domains}}          \\ \hline
       & \textbf{JTG-Synonym}     & \textbf{JTG-Skill} & \textbf{SCB-MT} & \textbf{XQuAD-r} \\ \hline
LaBSE  & 1.96               & 0.08           & 0.32            & 0.24             \\
BGE-m3 & 3.95               & 0.18           & 0.06            & 0.04             \\
mUSE   & 1.20               & 0.15           & 0.08            & \textbf{0.02}             \\
\textbf{Ours}   & \textbf{0.39}               & \textbf{0.08}           & \textbf{0.05}            & 0.04             \\ \hline
\end{tabular}%
}
\caption{Evaluation of language bias (LBKL $\downarrow$) for each sentence encoder on both in-domain and unseen domains.}
\label{tab:lan-bias-overall}
\end{table}

\begin{table}[ht]
\centering
\resizebox{\columnwidth}{!}{%
\Huge

\begin{tabular}{lc|ccc}
\hline
         & \textbf{in-domain} & \multicolumn{3}{c}{\textbf{Unseen-domains}}          \\ \hline
         & \textbf{JTG-Synonym}     & \textbf{JTG-Skill} & \textbf{SCB-MT} & \textbf{XQuAD-r} \\ \hline
mUSE     & 1.20               & 0.15           & 0.08            & 0.02             \\ \hline
+ (A) JT & 0.40               & 0.08           & 0.05            & 0.04             \\
+ (B) JD & 1.26               & 0.08           & 0.04            & 0.05             \\
+ (C) JF & 1.19               & 0.15           & 0.08            & 0.02             \\ \hline
\textbf{Ours}     & 0.39               & 0.08           & 0.05            & 0.04             \\ \hline
\end{tabular}%
}
\caption{A language bias comparison (LBKL $\downarrow$) each of our proposed training tasks on in-domain and unseen domains.}
\label{tab:abla-all}
\end{table}

\section{Conclusion}
In this paper, we introduced the bilingual sentence encoder for Thai-English for general-purpose use in the job recruitment domain. We employed a multi-task dual encoder framework that integrates three job-related tasks. Our evaluation further focused on both cross-lingual performance and language bias. To measure language bias, we proposed a novel metric called Language Bias Kullback–Leibler Divergence (LBKL), to quantify bias within the model. Our method consistently improves on both the synonym retrieval and job title classification tasks. Moreover, it demonstrated superior cross-lingual performance and greatly lowered the language bias compared to other state-of-the-art models.

\section{Acknowledgements}
This work is supported in part by JOBTOPGUN, a jobposting and recruitment platform in Thailand.


\appendix


{\small\bibliography{aaai25}}

\end{document}